\documentclass{sigchi}

\usepackage{balance}      
\usepackage{graphics}     
\usepackage{times}        
\usepackage{url}          

\makeatletter
\def\url@leostyle{%
  \@ifundefined{selectfont}{\def\UrlFont{\sf}}{\def\UrlFont{\small\bf\ttfamily}}}
\makeatother
\def\pprw{8.5in}
\def\pprh{11in}

\usepackage[pdftex]{hyperref}
\hypersetup{
pdftitle={SIGCHI Conference Proceedings Format},
pdfauthor={LaTeX},
pdfkeywords={SIGCHI, proceedings, archival format},
bookmarksnumbered,
pdfstartview={FitH},
colorlinks,
citecolor=black,
filecolor=black,
linkcolor=black,
urlcolor=black,
breaklinks=true,
}

\makeatletter
\def\@copyrightspace{\relax}
\makeatother

\begin{document}

\title{Predicting Daily Activities From Egocentric Images Using Deep Learning}

\numberofauthors{7}
\author{
  \alignauthor Daniel Castro*\\
    \email{dcastro9@gatech.edu}
  \alignauthor Steven Hickson*\\
    \email{shickson@gatech.edu}
  \alignauthor Vinay Bettadapura\\
    \email{vinay@gatech.edu}
  \alignauthor Edison Thomaz\\
    \email{ethomaz3@gatech.edu}
  \alignauthor Gregory Abowd\\
    \email{abowd@gatech.edu}
  \alignauthor Henrik Christensen\\
    \email{hic@cc.gatech.edu}
  \alignauthor Irfan Essa\\
    \email{irfan@cc.gatech.edu}\\ 
    \and
    \normalsize Georgia Institute of Technology\\
    \url{http://www.cc.gatech.edu/cpl/projects/dailyactivities}\\
    \small *These authors contributed equally to this work
} 

\maketitle

\begin{abstract}
We present a method to analyze images taken from a passive egocentric wearable camera along with the contextual information, such as time and day of week, to learn and predict  everyday activities of an individual. We collected a dataset of 40,103 egocentric images over a 6 month period with 19 activity classes and demonstrate the benefit of state-of-the-art deep learning techniques for learning and predicting daily activities. Classification is conducted using a Convolutional Neural Network (CNN) with a classification method we introduce called a late fusion ensemble. This late fusion ensemble incorporates relevant contextual information and increases our classification accuracy. Our technique achieves an overall accuracy of 83.07\% in predicting a person's activity across the 19 activity classes. We also demonstrate some promising results from two additional users by fine-tuning the classifier with one day of training data. \end{abstract}



\section{Introduction}

\begin{figure}[!h]
\centering
\includegraphics[width=0.9\columnwidth]{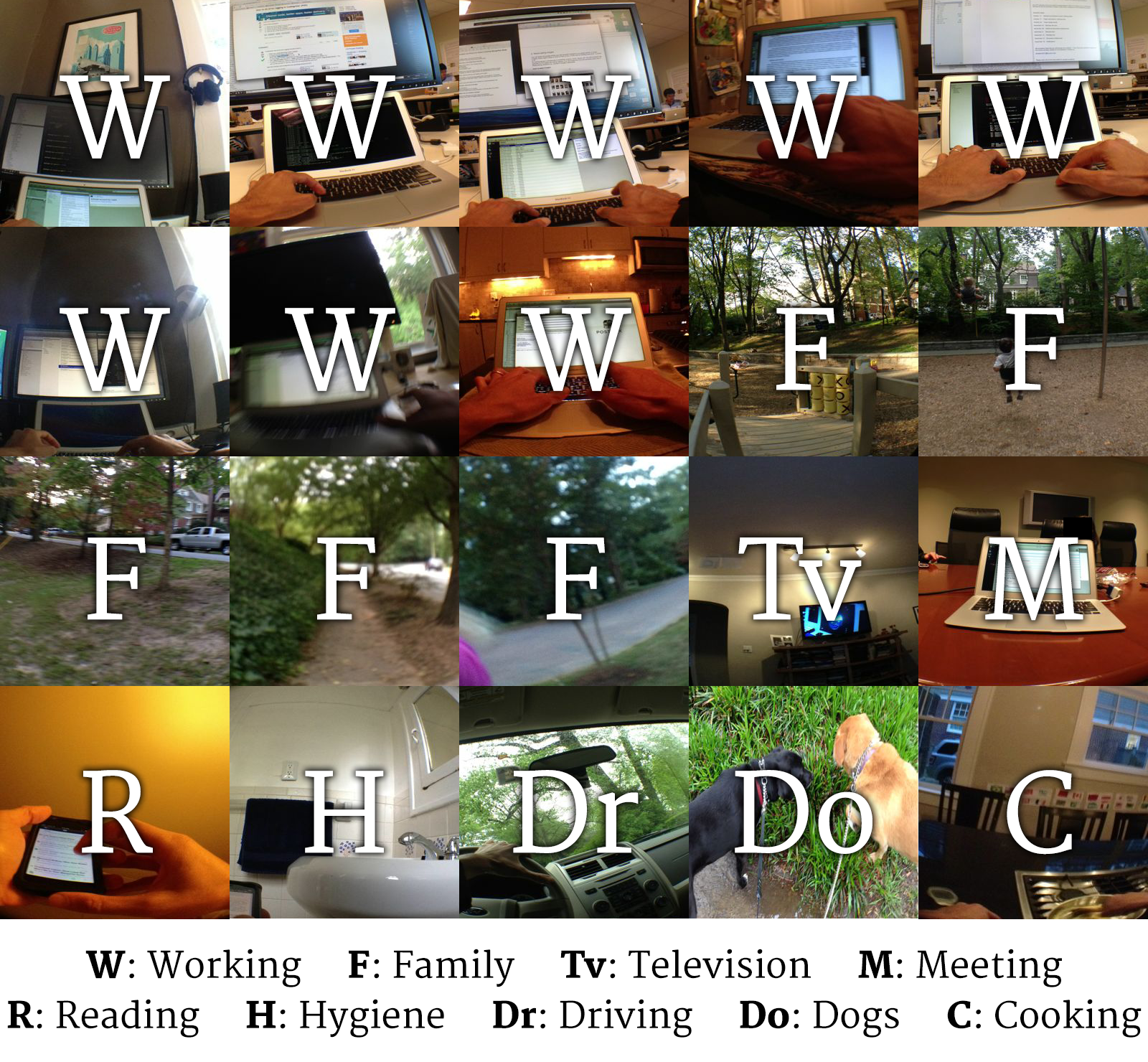}
\caption{Example images from our dataset of 40,000 egocentric images with their respective labels. The classes are representative of the number of images per class for the dataset. Note: We handpicked family images for this figure so they did not contain family subjects (for privacy and anonymity concerns).}   
\label{fig:input_output}
\end{figure}

The ability to automatically monitor and infer human behaviors in naturalistic environments is essential for a wide range of applications in areas such as context-aware personal assistance, healthcare, and energy management. Recently, wearable egocentric cameras such as the GoPro \footnote{http://www.gopro.com} and Narrative \footnote{http://www.getnarrative.com} have become ubiquitous, enabling a new form of capturing human experience. The first-person perspective photos taken by these cameras can provide rich and objective evidence of a person's everyday activities. As a result, this data collection approach has been extensively used in a variety of research domains, particularly in healthcare. Health-related applications that have leveraged first-person photos include, but are not limited to, retrospective memory support \cite{Hodges:2006uj}, dietary assessment \cite{OLoughlin:2013ic,Sun:2010gc}, autism support \cite{Marcu:2012wz}, travel and sedentary behavior assessment \cite{Kelly:2011ei,Kerr:2013dc}, and recognition of physical activities \cite{Zhang:2010ki}.

Besides the ever important issue of privacy, first-person photo capture with wearable cameras has one additional and serious challenge. Once photographs have been taken, in many cases, it is necessary to review them to identify moments and activities of interest, and possibly to remove privacy-sensitive images. This is particularly challenging when wearable cameras are programmed to take snapshots periodically, for example: every 30 or 60 seconds. At this rate, thousands of images are captured every week, making it imperative to automate and personalize the process of image analysis and categorization. 

We describe a computational method leveraging state-of-the-art methodologies in machine learning to automatically learn a person's behavioral routines and predict daily activities from first-person photos and contextual metadata such as day of the week and time. Example of daily activities include cooking, eating, watching TV, working, spending time with family, and driving (see Table~\ref{tab:distribution} for a full list). The ability to objectively track such daily activities and lifestyle behaviors is extremely valuable since behavioral choices have strong links to chronic diseases \cite{Willett:2002jy}. 

To test and evaluate our method, we compiled a dataset of 40,103 images representing everyday human activities. The dataset has 19 categories of activities and were collected by one individual over a period of six months ``in the wild". Given the egocentric image and the contextual date-time information, our method achieves an overall accuracy of 83.07\% at determining which one of these 19 activities the user is performing at any moment. 

Our classification method uses a combination of a Convolutional Neural Network (CNN) and a Random Decision Forest (RDF), using what we refer to as a CNN late-fusion ensemble. It is designed to work on single images captured over a regular interval as opposed to video clips. Capturing hours of egocentric video footage would require tethered power and large storage bandwidth, which still remains impractical. An example of our input egocentric image and the output class prediction probabilities is shown in Figure \ref{fig:input_output}. In brief, our contributions are:

\begin{itemize}
    \item A robust framework for the collection and annotation of egocentric images of daily activities from a wearable camera. 
    \item A CNN+RDF late-fusion ensemble that reduces overfitting and allows for the inclusion of local image features, global image features, and contextual metadata such as day of the week and time.
    \item A promising approach to generalize and fine-tune the trained model to other users with a minimal amount of data and annotation by the user. We also get insights into the amount of data the first user needs to collect to train a classifier and how much data subsequent users need to collect to fine-tune the classifier to their lifestyle.
    \item A unique dataset of annotated egocentric images spanning a 6 month period and a CNN+RDF late-fusion ensemble model fit to that data.
 
\end{itemize}

\section{Related Work}

\textbf{Activity Analysis:} Discovering daily routines in human behavior from sensor data has been an active area of research. With a dataset of 46 days of GPS sensor data collected from 30 volunteer subjects, Biagioni and Krumm demonstrated an algorithm that uses location traces to assess the similarity of a person's days \cite{Biagioni:2013tg}. Blanke and Schiele explored the recognition of daily routines through low-level activity spotting, with precision and recall results in the range of 80\% to 90\% \cite{Blanke:2009tq}. Other proposed techniques for human activity discovery have included non-parametric approaches \cite{Sun:2014jk}, and topic modeling \cite{Huynh:2008tl}.

One of the most comprehensive computer-mediated analysis of human behaviors in naturalistic settings was done by Eagle and Pentland~\cite{Eagle:2006vw}. By collecting data from 100 mobile phones over a 9-month period, they were able to recognize social patterns in daily user activity, infer relationships, identify socially significant locations, and model organizational rhythms. Their work was based on a formulation for identifying structure in routine called \textit{eigenbehaviors} \cite{Eagle:2009jx}. By examining a weighted sum of an individual's eigenbehaviors, the researchers were able to predict behaviors with up to 79\% accuracy. This approach also made it possible to calculate similarities between groups of individuals in terms of their everyday routines. With data collected in-the-wild over 100 days, Clarkson also presented an approach for the discovery and prediction of daily patterns from sensor signals \cite{Clarkson:2002vh}.

While long-term activity prediction approaches have mostly relied on mobile phone data and sensor signals, our approach is focused on the prediction of human activities in real-wold setting from first-person egocentric images using computer vision and machine learning approaches. While there has been some work on detecting activities with egocentric cameras, most of these approaches rely on video and hand-crafted features. Fathi et al. \cite{fathi2011understanding} used egocentric video and detected hands and objects to recognize actions. Pirsiavash et al. \cite{pirsiavash2012detecting} introduced an annotated dataset that includes 1 million frames of 10 hours of video collected from 20 individuals performing activities of daily living in 20 different homes and used hand-crafted object detectors and spatial pyramids to classify activities using a SVM classifier.

In contrast to state-of-the-art approaches that use hand-crafted features with traditional classification approaches on egocentric images and videos, our approach is based on Convolutional Neural Networks (CNNs) combining image pixel data, contextual metadata (time) and global image features. Convolutional Neural Networks have recently been used with success on single image classification with a vast number of classes \cite{krizhevsky2012imagenet} and have been effective at learning hierarchies of features \cite{zeiler2014visualizing}. However, little work has been done on classifying activities on single images from an egocentric device over extended periods of time. This work aims to explore that area.

\textbf{Privacy Concerns:} One of the challenges of continuous and automatic capture of first person point-of-view images is that these images may, in some circumstances, pose a privacy concern. Privacy is an area that deserves special attention when dealing with wearable cameras, particularly in public settings. Kelly et al. proposed an ethical framework to formalize privacy protection when wearable cameras are used in health behavior research and beyond \cite{Kelly:2013ee} while Thomaz et al. proposed a framework for understanding the balance between saliency and privacy when examining images, with a particular focus on photos taken with wearable cameras \cite{Thomaz:2013iv}. People's perceptions of wearable cameras are also very relevant. Nguyen et al. examined how individuals perceive and react to being recorded by a wearable camera in real-life situations \cite{Nguyen:2009tv}, and Hoyle et al. studied how individuals manage privacy while capturing lifelong photos with wearable cameras \cite{Hoyle:2014dj}.

\section{Data Collection}

Over a period of 26 weeks, we collected 40,103 egocentric images of activities of daily living for one subject with 19 different activity classes. The images were annotated manually using a tool we developed to facilitate this arduous daily task. The classes were generated by the subject at their discretion based on what activities the user conducted (we did not provide the labels prior to data collection).

\subsection{Process}

\begin{figure}[!t]
\centering
\includegraphics[width=0.6\columnwidth]{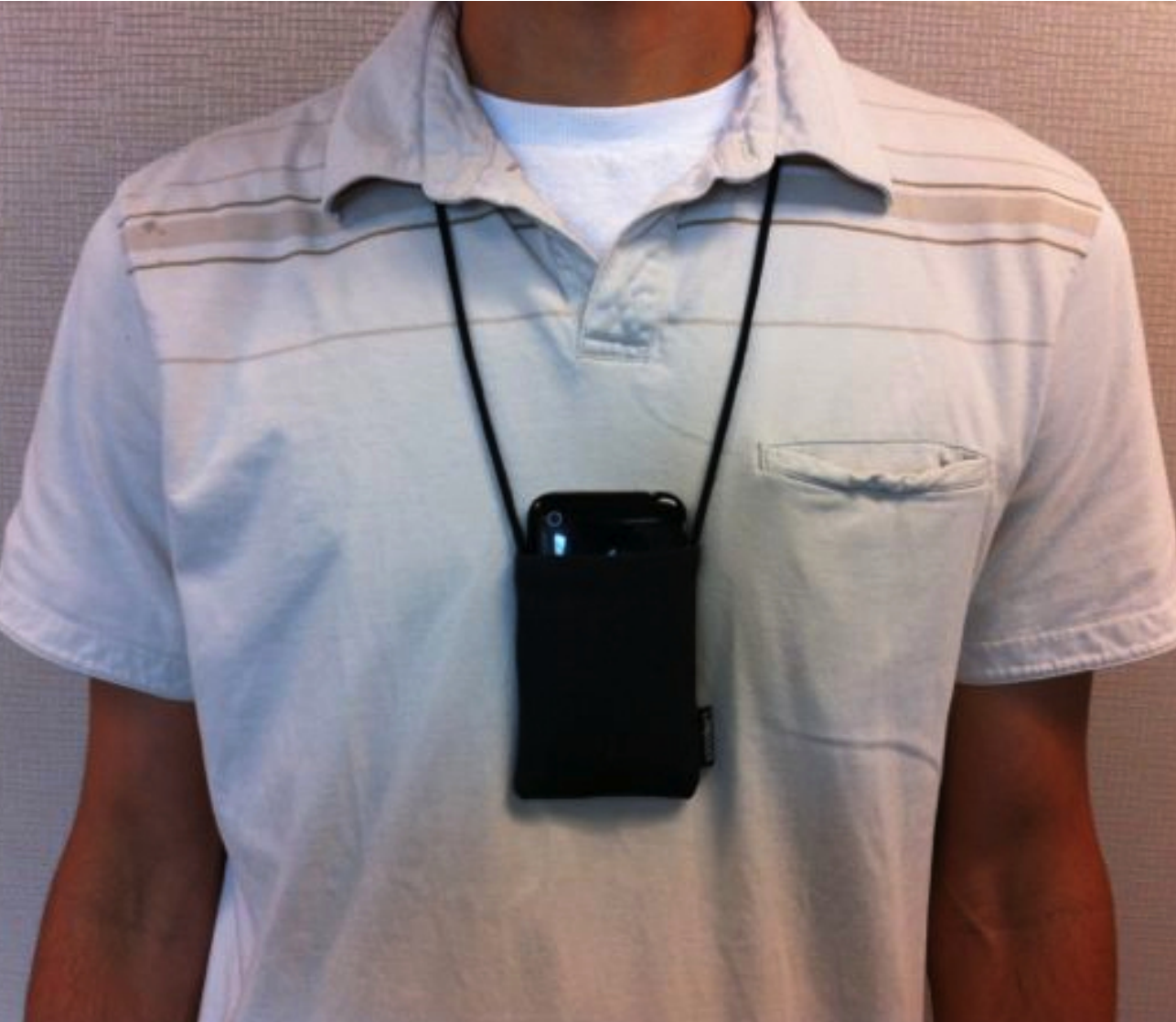}
\caption{Our simple yet practical setup for data collection: A neck identity holder fitted with a smartphone with our custom data collection application. No specialized or expensive hardware is needed.}
\label{fig:setup_img}
\end{figure}

The subject was equipped with a neck identity holder that was fitted to hold a smartphone in portrait mode (shown in Figure \ref{fig:setup_img}). We developed an application that runs on the smartphone and captures photos at fixed intervals, which allows for the capture of egocentric data throughout the day. At the end of the day, the participant could filter through the images in order to remove unwanted and privacy sensitive images and annotate the remaining images. The participant categorized the data collected for 26 weeks using the annotation tool described in the following subsection into one of the 19 activity classes. The distribution of these classes is shown in Table \ref{tab:distribution}.  We can see that "Working" and "Family" are the top two dominant classes due to the participant's lifestyle. We note that the participant was free to collect and annotate data at their disclosure. The subject was also free to leave ambiguous images (i.e. going from work to a meeting) unannotated. Any unlabeled and deleted images were reasonably not included in the dataset.

\subsection{Tool for Annotation}
\label{sec:annotation}

We developed a tool for rapid image annotation that is intended for daily activity labeling. The tool automatically receives the imagery taken from the application on the egocentric device and displays them in chronological order. The user is then able to select sequential images (in chunks) to label as specific activities. This facilitates the process of labeling large image sets in a simpler and intuitive manner.

\subsection{Description of Dataset}

As shown in Table \ref{tab:distribution}, the distribution of tasks is represented by a few common daily tasks followed by semi-frequent activities with fewer instances. We are keen to highlight the difficulty of certain classes due to their inherent overlap (socializing vs. chatting, chores vs. family, cleaning vs. cooking, etc). This class overlap is due to the inherent impossibility of describing a specific moment with one label (the participant could be eating and socializing). 

\begin{table}[t]
\begin{center}\begin{small} 
\begin{tabular}{|l|c|c|}
\hline \textbf{Classes} & \textbf{Number of Images} & \textbf{Percent of Dataset} \\ \hline \hline
Chores & 725 & 1.79 \\ \hline
Driving & 1031 & 2.54 \\ \hline
Cooking & 759 & 1.87 \\ \hline
Exercising & 502 & 1.24 \\ \hline
Reading &  1414 & 3.48 \\ \hline
Presentation & 848 & 2.09 \\ \hline
Dogs & 1149 & 2.83 \\ \hline
Resting & 106 & 0.26 \\ \hline
Eating & 4699 & 11.58 \\ \hline
Working & 13895 & 34.24 \\ \hline
Chatting & 113 & 0.28 \\ \hline
TV & 1584 & 3.90 \\ \hline
Meeting & 1312 & 3.23 \\ \hline
Cleaning & 642 & 1.59 \\ \hline
Socializing & 970 & 2.39 \\ \hline
Shopping & 606 & 1.49 \\ \hline
Biking & 696 & 1.71 \\ \hline
Family & 8267 & 20.37 \\ \hline
Hygiene & 1266 & 3.12 \\ \hline
\end{tabular}
\end{small}
\end{center}
\caption{The distribution of the 19 different classes in our dataset.}
\label{tab:distribution}
\end{table}

\begin{figure*}[t]
\begin{center}
   \includegraphics[width=\linewidth]{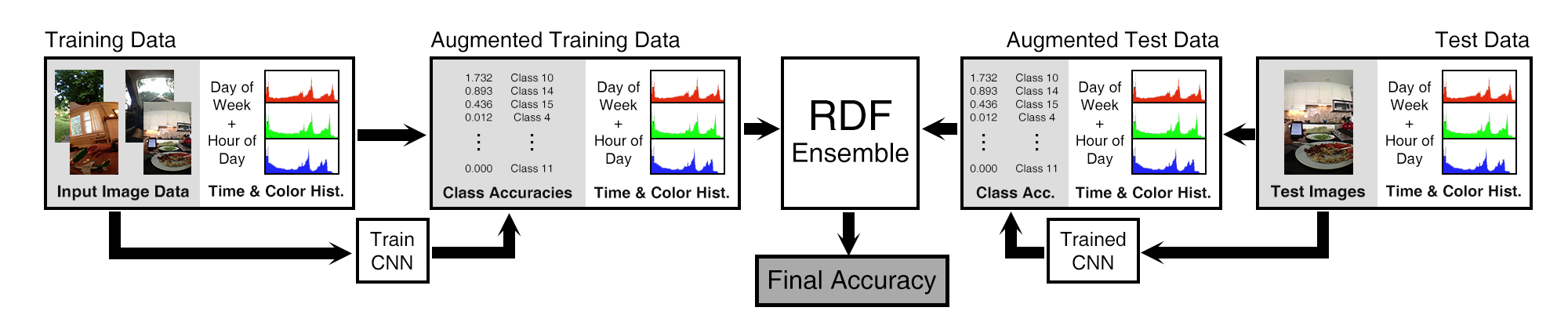}
\end{center}
   \caption{Overview of our Convolutional Neural Network Late Fusion Ensemble for predicting activities of daily living.}
\label{fig:cnn_lfe}
\end{figure*}

The bi-weekly breakdown of data collection is shown in Table \ref{tab:time_dist}. We can see a general increase in the number of annotated samples later in the collection process. Some of this is due to increasing the interval at which the application captured images up to once a minute from once every five minutes. The rest of the increase can be attributed to the participant becoming more comfortable with the data collection and annotation process, and over time, successfully incorporating this process into their day-to-day routine.

The participant collected the majority of the data from approximately 7-8am to 7-8pm. The majority of the data that is not captured is therefore during the participants sleep cycle. On an average day we retain 80\% of the data that is collected (the participant removes approximately 20\% for privacy and null classes). The participant handled null classes (blurry images, etc) by leaving them unlabeled. These images were then removed prior to assembling the dataset.

\begin{table}[t]
\begin{small} 
\begin{center}
\begin{tabular}{|l|c|c|}
\hline \textbf{Classes} & \textbf{Number of Samples} & \textbf{Percent of Dataset} \\ \hline \hline
Week 1\&2 & 553 & 1.40 \\ \hline
Week 3\&4 & 814 & 2.07 \\ \hline
Week 5\&6 & 69 & 0.18 \\ \hline
Week 7\&8 & 216 & 0.55 \\ \hline
Week 9\&10 & 239 & 0.61 \\ \hline
Week 11\&12 & 2586 & 6.58 \\ \hline
Week 13\&14 & 5858 & 14.90 \\ \hline
Week 15\&16 & 6268 & 15.94 \\ \hline
Week 17\&18 & 2903 & 7.38 \\ \hline
Week 19\&20 & 3417 & 8.69 \\ \hline
Week 21\&22 & 6465 & 16.45 \\ \hline
Week 23\&24 & 4695 & 11.94 \\ \hline
Week 25\&26 & 5229 & 13.30 \\ \hline
\end{tabular}
\end{center}
\end{small}
\caption{The bi-weekly distribution of the number of images in our dataset.}
\label{tab:time_dist}
\end{table}

\section{Methodology}

We present a methodology for incorporating contextual metadata and other traditional hand-crafted features with a Convolutional Neural Network (CNN) that processes the raw image data. The method is compared to baseline machine learning methods (k-Nearest Neighbors (kNN) \cite{cover1967nearest} and Random Decision Forests (RDF) \cite{breiman2001random}) in order to demonstrate the benefits and limitations of our approach. We also introduce a method called late fusion ensembling for combining non-image data with CNN probabilities and compare it to a traditional CNN and classic ensembling methods. 

\subsection{Baseline Approaches}

We ran evaluations using k-Nearest Neighbor (kNN) and Random Decision Forest (RDF) classifiers in order to adequately fine-tune the best accuracy for our baseline. We parametrized our dataset using contextual metadata (day of the week (as a nominal value from 0 to 6) and time of day) and global image features (color histograms). We found that a kNN classifier (with a k-value of 3) trained on the metadata and the color histograms (with 10 bins) gave an accuracy of 73.07\% which was better than training a kNN trained on the metadata alone or the color histograms alone. We tested the classifier at incremental parameters of k (until 50) and found that performance slowly degraded as we increased k beyond 3. We further tested the time metadata at three granularities (the hour, hour + minutes (i.e. 7:30am = 7.5), and hour and minute as separate features) and found the difference in prediction accuracy to be negligible due to the scheduled nature of humans. We selected to keep the hour and minute as separate features as it had the highest accuracy. Further, we found that a RDF classifier with 500 trees trained on the metadata and color histograms (with 10 bins) gave us the best overall accuracy of 76.06\% (note that random chance, by picking the highest prior probability, is 34.24\% for this dataset). Training the RDF with more than 500 trees had a negligible effect on the total accuracy. Our baseline results can be seen in Table \ref{tab:baseline}. It is important to note that a high total accuracy is driven by the distribution of the data amongst the classes. Since a majority of the data is in two classes (``Working" and ``Family"), a classifier can achieve a high total accuracy by accurately classifying only those two classes. We also show average class accuracy to show how well the baseline classifier does for all classes distributed evenly.

\subsection{Convolutional Neural Network}

\begin{figure}[t]
\begin{center}
   \includegraphics[width=0.95\linewidth]{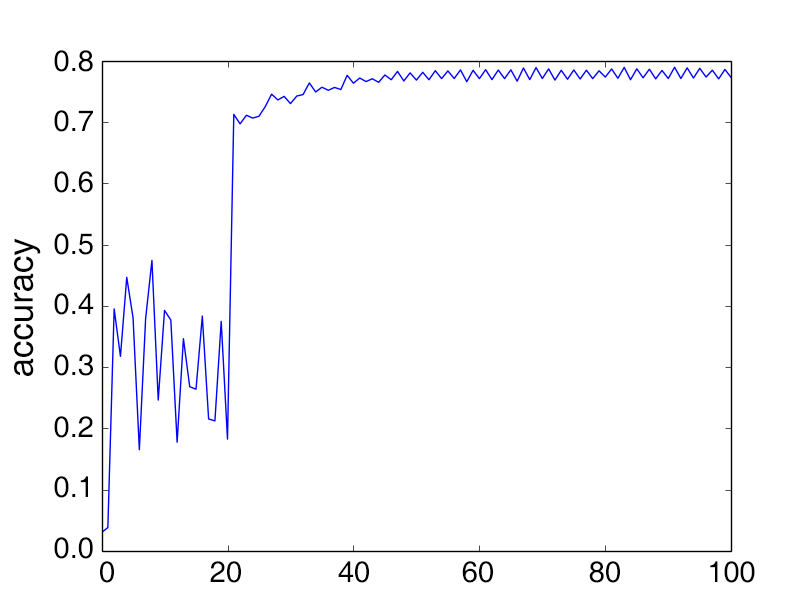}
\end{center}
   \caption{A Convolutional Neural Network trained for 100,000 iterations. We can see the accuracy convergence after 20,000 to 30,000 iterations.}
\label{fig:iterations}
\end{figure}

Recently, Convolutional Neural Networks (CNNs)\cite{lecun1998gradient} have been shown to be effective at modeling and understanding image content for classification of images into distinct, pre-trained classes. We used the Caffe CNN framework \cite{jia2014caffe} to build our model since it has achieved good results in the past and has a large open-source community. Since the dataset has a small number of images, we fine-tune our CNN using the methodology of \cite{hinton2012improving} using the ImageNet \cite{deng2009imagenet} classification model introduced by Krizhevsky et al. in \cite{krizhevsky2012imagenet} that was trained on over a million images in-the-wild. We retrain the last layer using our collected data with 19 labels for daily activity recognition. We set the base learning rate to 0.0001 in order to converge with our added data and use the same momentum of 0.9 and weight decay of 0.0005 as \cite{krizhevsky2012imagenet} with up to 100,000 iterations as shown in Figure \ref{fig:iterations}. Our CNN has five convolutional layers, some max-pooling layers, and three fully-connected layers followed by dropout regularization and a softmax layer with an image size of 256x256 just as in \cite{krizhevsky2012imagenet}. We split our data by classes into 75\% training, 5\% validation, and 20\% testing. The classifier was never trained with testing data on any of the experiments. The parameters were chosen using the validation set and the fine tuning in \b{all} of the experiments was only done with the training set. It is interesting to note that the algorithm jumps to almost ~78\% accuracy after only 20,000 to 30,000 iterations and converges around 50,000 iterations due to fine tuning. Despite a high total accuracy, the class accuracy of a CNN alone is hindered due to the lack of contextual information and global image cues.

For many problems with small amounts of data, data augmentation can be effective at preventing overfitting and increasing accuracy. However, in this case, we are collecting data at a specific orientation and viewpoint, so many data collection techniques are not applicable. Because of this, we elected not to augment our training data although that would be a useful extension of the work.

\begin{table*}[t]
\begin{small} \begin{center}
\begin{tabular}{|l|l|l|l|l|l|l|}
\hline & \textbf{kNN Metadata} & \textbf{kNN Hist} & \textbf{kNN Metadata+Hist} & \textbf{RDF Metadata} & \textbf{RDF Hist} & \textbf{RDF Metadata+Hist} \\ \hline \hline
Avg. Class Accuracy &  15.51 &  44.23 &  54.72 &  15.51 &  40.43 &  50.71 \\ \hline
Total Accuracy      &  52.50 &  65.62 &  73.07 &  52.50 &  68.89 &  76.06 \\ \hline
\end{tabular}
\caption{A comparison of the baselines using kNN and RDF trained on contextual metadata, color histograms and a combination of both.}
\label{tab:baseline}
\end{center}
\end{small}
\end{table*}

\subsection{Classic Ensemble}
One method to combine the CNN output with non-image data is a classic ensemble method. Training a classifier such as a RDF on the contextual metadata can yield a probability distribution which can be combined with the CNN probability distribution to yield a final probability. This equally weights the CNN output and the RDF output in order to get the best output possible. This can prevent over-fitting from the CNN but doesn't necessarily increase the prediction accuracies since it doesn't leverage which classifier is better at which classes or which information from the classifiers are important. 

\subsection{Late Fusion Ensemble}
To solve the problem of combining a CNN with a classic ensemble, we developed a late-ensemble technique. We use a RDF trained on the CNN soft-max probabilities along with the contextual metadata (day of week and time of day) and the global image information (histograms of color), each being separate features for the RDF. This allows for a good combination of outputs that can be learned rather than naively combined. Using this we outperform the classic ensemble and the normal CNN model by approximately 5\%. The pipeline for our method is shown in Figure \ref{fig:cnn_lfe}.

\section{Results}

In this section we present a comparison of baseline machine learning techniques against the different convolutional approaches for the classification of daily living activities. As shown in Table \ref{tab:baseline}, kNN and RDF perform surprisingly well with contextual metadata (day of the week and time of day) and color histograms. RDFs marginally outperform the kNN methods, particularly with the use of color histograms. It is worth mentioning that we tested other global features (such as GIST \cite{oliva2001modeling}) on the same baseline methods and obtained negligible changes in accuracy.

\begin{figure}[!h]
\centering
   \includegraphics[width=\linewidth]{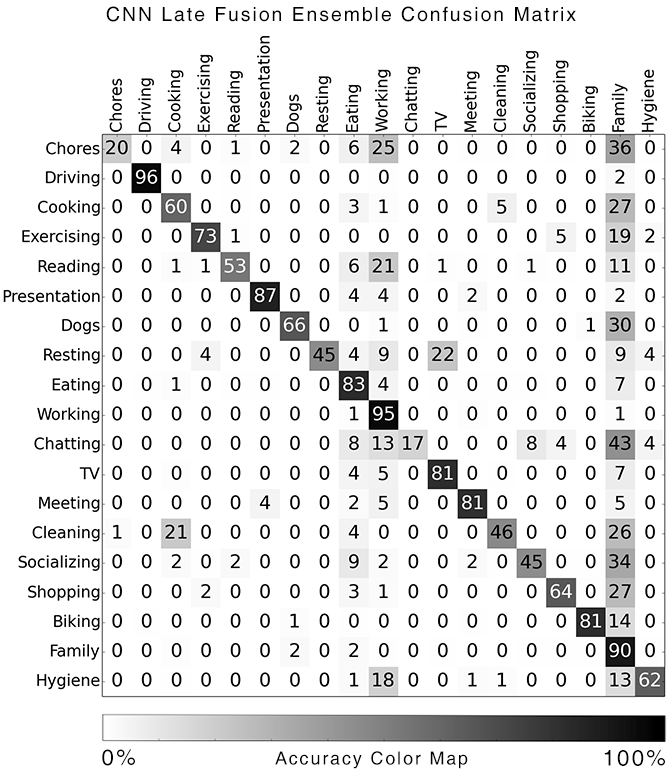}          
\caption{Confusion Matrix for the 19 classes of our dataset with columns as the predicted labels and rows as the actual labels.}
\label{fig:conf}
\end{figure}

In order to improve the performance of our activity prediction we leverage the use of local image information. Through the use of a regular CNN, we see a minor increase in total accuracy (+2\%) over the baseline (see Table \ref{tab:cnn}), but a much more impressive jump in average class accuracy (+7\%). We see an even greater increase in accuracy as we incorporate both contextual metadata and global image information (color histograms). We have demonstrated through the baseline methods that these features are of importance, which is why we developed our CNN late fusion ensemble that leverages the metadata and global and local image features. Our best ensemble leverages all of the presented information for a total accuracy of 83.07\% with an average class accuracy of 65.87\% showing an impressive increase over the baseline and the other methods. A confusion matrix of our final method's results is shown in Figure \ref{fig:conf}.

\begin{table}
\begin{small} \begin{center}
\begin{tabular}{|l|c|c|c|c|}
\hline                  &            \textbf{kNN} &            \textbf{RDF} &            \textbf{CNN} & \textbf{CNN+LF} \\
\hline \hline
Chores                  & \textbf{33.10} &          17.24 &          00.69 &          20.00              \\ \hline
Driving                 &          55.07 &          60.87 & \textbf{98.55} &          96.62              \\ \hline
Cooking                 &          25.66 &          35.53 &          47.37 & \textbf{60.53}              \\ \hline
Exercising              &          44.00 &          63.00 &          69.00 & \textbf{73.00}              \\ \hline
Reading                 & \textbf{68.55} &          49.12 &          30.04 &          53.36              \\ \hline
Presentation            &          80.00 &          72.35 &          80.59 & \textbf{87.06}              \\ \hline
Dogs                    &          62.17 &          44.35 &          55.65 & \textbf{66.09}              \\ \hline
Resting                 & \textbf{72.73} &          54.55 &          27.27 &          45.45              \\ \hline
Eating                  &          77.14 &          75.75 &          82.05 & \textbf{83.12}              \\ \hline
Working                 &          91.10 & \textbf{96.42} &          93.49 &          95.19              \\ \hline
Chatting                & \textbf{21.74} &          04.35 &          00.00 &          17.39              \\ \hline
TV                      &          77.38 &          75.79 & \textbf{81.75} & \textbf{81.75}              \\ \hline
Meeting                 &          68.73 &          61.00 &          73.36 & \textbf{81.47}              \\ \hline
Cleaning                &          26.56 &          30.47 &          38.28 & \textbf{46.09}              \\ \hline
Socializing             & \textbf{52.85} &          37.31 &          31.60 &          45.08              \\ \hline
Shopping                &          40.16 &          27.87 &          63.93 & \textbf{64.75}              \\ \hline
Biking                  &          19.57 &          23.19 &          78.26 & \textbf{81.88}              \\ \hline
Family                  &          70.82 &          87.42 &          86.69 & \textbf{90.15}              \\ \hline
Hygiene                 &          52.36 &          46.85 &          51.57 & \textbf{62.60}              \\ \hline
Avg. Class Accuracy     &          54.72 &          50.71 &          57.38 & \textbf{65.87}              \\ \hline
\textbf{Total Accuracy}          &          73.07 &          76.06 &          78.56 & \textbf{83.07}              \\ \hline
\end{tabular}
\end{center}
\end{small}
\caption{A comparison of the best of all methods (using contextual metadata, color histograms and pixel data) for all the 19 activity classes. CNN+LF is CNN with Late Fusion Ensemble}
\label{tab:results}
\end{table}

\begin{table*}[t]
\begin{center}
\begin{tabular}{|l|l|l|}
\hline & \textbf{Average Class Accuracy} & \textbf{Total Accuracy} \\ \hline \hline
CNN                                                & 57.38 & 78.56 \\ \hline
CNN Classic Ensemble (Pixel + Metadata)            & 53.48 & 78.47 \\ \hline
CNN Classic Ensemble (Pixel + Metadata + Hist)            & 59.72 & 81.49 \\ \hline
CNN Late Fusion Ensemble (Pixel)                   & 63.22 & 80.94 \\ \hline
CNN Late Fusion Ensemble (Pixel + Metadata)        & 65.29 & 82.45 \\ \hline
CNN Late Fusion Ensemble (Pixel + Metadata + Hist) & \textbf{65.87} & \textbf{83.07} \\ \hline
\end{tabular}
\caption{A comparison of different CNNs and CNN ensembles using contextual  metadata, global features (color histograms), raw image pixels and their combinations.}
\label{tab:cnn}
\end{center}
\end{table*}

\section{Discussion}

Our method achieves the highest accuracy on the classes with the most samples (as one would expect since test accuracy increases with larger amounts of training data). As shown in Table \ref{tab:cnn}, our ensemble method outperforms both a normal CNN and a classic ensemble with a CNN. Training an RDF with extra features and the CNN probabilities allows the RDF to find what is important for each individual class. It also allows for the other types of data to be effectively added in a framework that prevents some of the overfitting that CNNs typically have. This shows how our novel ensemble method effectively combines local pixel-level information, contextual information, and global image-level information. Because it relies on a CNN running on a GPU, the system uses a large amount of power and is not well suited for embedded devices. On an ARM device, testing each image would take more than 15 seconds. However, the method could be run on a server that an embedded device could query.

\begin{figure}[!h]
\centering
   \includegraphics[width=\columnwidth]{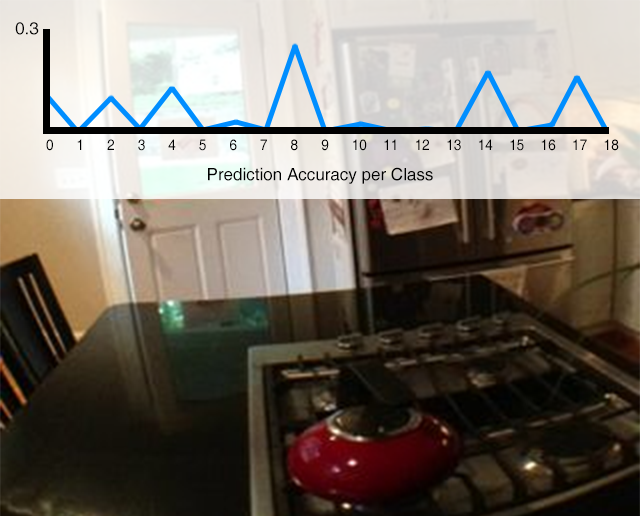}
\caption{An example of a classification error on an image from the class ``Chores" (class 0).  We can see the confusion against ``Eating" (class 8), ``Socializing" (class 14) and ``Family" (class 17) due to the presence of the kitchen environment in the image.}
\label{fig:error}
\end{figure}

Many of the classification failures of our method deal with some classes being inter-related. Our worst results are in ``Chores" and ``Chatting". These classes can be easily confused with others such as ``Cleaning", ``Working" and ``Family". In many examples in which the subject is conducting a chore, the family is in the background, which may confuse the classifier. An example of a chore misclassification can be seen in Figure \ref{fig:error}. In this example, the image has erroneous probability peaks for ``Eating", ``Socializing" and ``Family" classes due to the presence of the kitchen environment in the image, a place where the family meet, socialize and eat together. We acknowledge this as a limitation of the method used for data capture that uses a single image frame in contrast to a short video clip. We believe the extension of our method to short video clips would prevent some of these difficult classification errors but would present further questions in privacy, device storage and battery life.

\begin{figure*}[t]
\begin{center}
\includegraphics[width=0.8\linewidth]{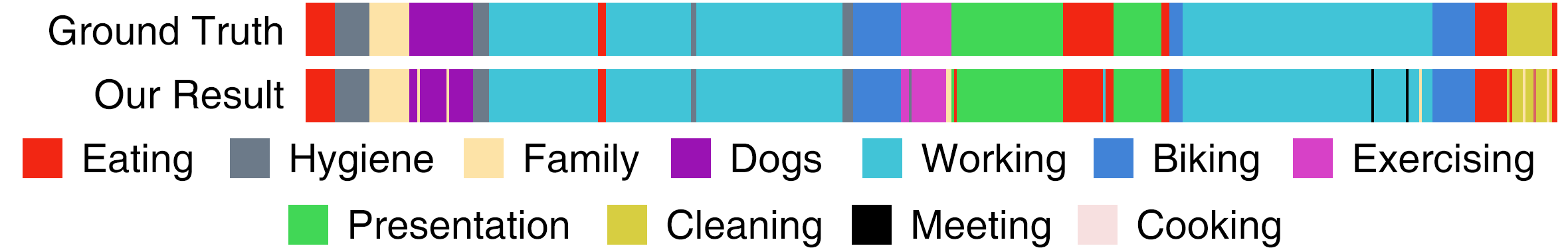}
\end{center}
\caption{An example of a randomly chosen day and the classifier's predicted output.}
\label{fig:dailyExample}
\end{figure*}

To visually display an average day and our prediction of the activities for that day, we have taken a random daily sample from the data and classified it visually. The results are shown in Figure \ref{fig:dailyExample}. For this particular day, our classifier nears 100\% accuracy at predicting the user's activities. We are keen to highlight the misclassification errors on this given day. During the classification of ``Dogs" at the beginning of the day (seen in purple), we notice two slivers of misclassification in which the algorithm detects ``Family" instead of walking the dogs (both classes have instances of green foliage). We see similar errors in the last light blue segment, representing ``Working", in which it detects two instances as ``Meeting" instead of ``Working". This provides further evidence that the class overlap is likely to contribute heavily to the 16.93\% overall misclassification that we have in our dataset.

\begin{figure}[t]
\begin{center}
\includegraphics[width=0.85\linewidth]{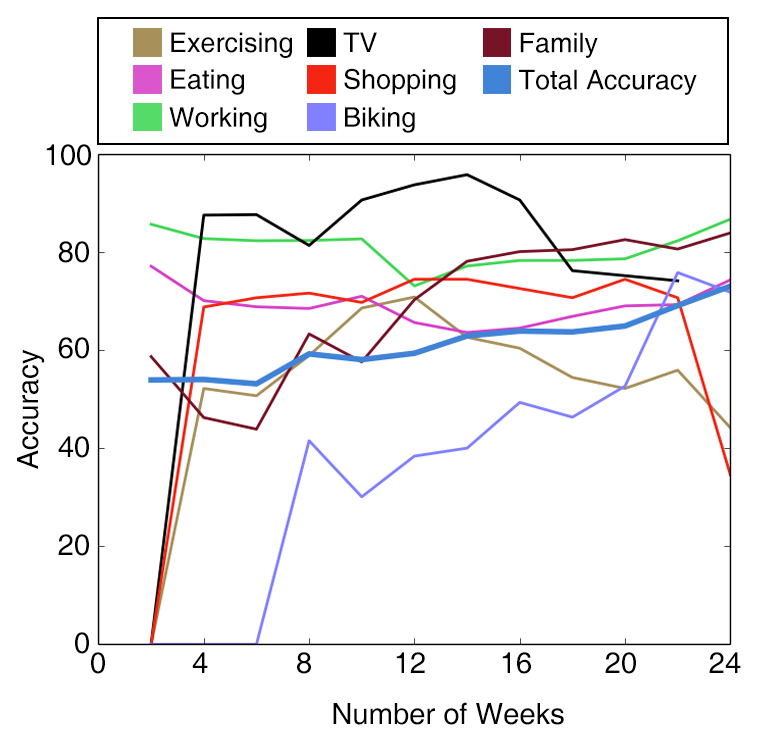}
\end{center}
\caption{A plot of class accuracies vs. the number of weeks of training samples. We can see a general trend where the class accuracies increase as the amount of training samples increase. A significant increase in accuracy is seen after training on the first 4 weeks of data.}
\label{fig:timeGraph}
\end{figure}

In a second experiment, we demonstrate the correlation between the amount of training data and the algorithms' test accuracy for the participant. We highlight two hypotheses for the increase in accuracy over time. The first is that the algorithm is adequately learning the participants' schedule and frequented activities, which allows it to better model their daily activities. The second plausible hypothesis is that the algorithm is adapting to general human behavior and learning the overall characteristics of specific classes. This presents two interesting questions for the applications of this research. First, how much data is required to train a generic model and second, how much data is required to ``fine-tune" said generic model to a specific user. We have tried to address the first of these questions by training our model with varying amounts of data points to observe the number of days/samples a user is required to collect in order to train a good generic model. The top 7 classes are shown in Figure \ref{fig:timeGraph} (plots for the other 12 classes are omitted to maintain clarity). We can see that the class accuracies improve as more data is captured with a significant increase in accuracy after the first 4 weeks.

In order to address the second question, we performed a final experiment in which two volunteers (V1 and V2) wore the egocentric device for 48 hours in order to collect images and time-stamps at a 60 second interval. The data was divided equally into a training and test set (Day 1 for training and Day 2 for testing) in order to test the validity of the model trained by our original participant's data. The results of this experiment are demonstrated in Table \ref{tab:gen}. As you can see, for some classes that involve a similar viewpoint and environment, like reading, the model generalizes very well. However, for many others such as driving and chatting where volunteers are going different places and talking to different people, the model does not generalize well. It is worth noting that the initial accuracy prior to fine-tuning performs worse than the highest prior probability of the original model (34.24\%). We reason that this is due to the difference in habits between participants (we work, read, cook for distinct periods of time) that require fine-tuning to adapt to one's specific daily schedule.

Different individuals also have different activities and one set of class labels from one individual might not fit another individual's lifestyle. Given the model trained for one person, is it possible to fine-tune the classifier to yield good results for a different person, even with different classes? At its core, this addresses the question of whether a classifier is learning the schedule and habits of one person or if the learning is inherently adapting to common human behavior. As seen in Table \ref{tab:gen}, the classifier trained on the original participant was not very successful. However, fine-tuning that model with just one day of data from the new user can yield very good accuracy. Not only did this achieve great accuracy, but the CNN converged in less than 5,000 iterations, whereas the original CNN takes more than 50,000 iterations to converge. This implies that part of the model is learning human behavior while another part is learning the habits of a specific person. We can use a small amount of training data to fine-tune the classifier to learn the habits of a new person, while still keeping the knowledge of general human behavior.

\begin{table}[t]
\begin{small}
\begin{center}
\begin{tabular}{|l|l|l|l|l|l|}
\hline & \textbf{Original} & \textbf{V1} & \textbf{V1 Fine} & \textbf{V2} & \textbf{V2 Fine} \\ \hline \hline
Chores                  & 20.00 & 5.56  & 25.0 & N/A & N/A            \\ \hline
Driving                 &  96.62 & 18.6   & 100.0 & 0.0 & 100.0             \\ \hline
Cooking                 &  60.53 & 0.0  & 25.0 & N/A & N/A              \\ \hline
Exercising              & 73.00 & 0.0  & 50.0 & N/A & N/A              \\ \hline
Reading                 &  53.36 & 77.78  & 75.0 & N/A & N/A              \\ \hline
Presentation            & 87.06 & N/A  & N/A & N/A & N/A              \\ \hline
Dogs                    & 66.09 & N/A  & N/A & N/A & N/A              \\ \hline
Resting                 & 45.45 & N/A  & N/A & N/A & N/A              \\ \hline
Eating                  & 83.12 & 11.48  & 76.92 & 30.68 & 100.0             \\ \hline
Working                 &  95.19 & 31.59  & 98.32 & 39.14 & 94.44              \\ \hline
Chatting                & 17.39 & 0.0 & 86.67 & 0.0 & 96.72              \\ \hline
TV                      & 81.75 & 0.0  & 33.33 & N/A & N/A              \\ \hline
Meeting                 & 81.47 & 0.0  & 100.0 & 0.0 & 60.0              \\ \hline
Cleaning                & 46.09 & 0.0  & 0.0 & N/A & N/A              \\ \hline
Socializing             & 45.08 & 0.0  & 0.0 & 0.0 & 83.33              \\ \hline
Shopping                & 64.75 & 40.0  & 50.0 & N/A & N/A              \\ \hline
Biking                  & 81.88 & N/A  & N/A & N/A & N/A             \\ \hline
Walking                  & N/A & 0.0  & 57.14 & N/A & N/A             \\ \hline
Family                  & 90.15 & N/A  & N/A & N/A & N/A              \\ \hline
Hygiene                 & 62.60 & 13.33  & 0.0 & 27.78 & 81.82              \\ \hline
Class Acc    & 65.87 & 10.56  & 51.83 & 13.94 & 88.05              \\ \hline
Total Acc          & 83.07 & 23.58  & 86.76 & 27.06 & 91.23              \\ \hline
\end{tabular}
\caption{A comparison of the original model tested on two volunteers and the fine tuned model. ``Original" is the original applicants data and model. ``V1" and ``V2" are the results from the original model tested on volunteers 1 and 2 data respectively. ``V1 Fine" and ``V2 Fine" are the results from the fine-tuned models trained on volunteers 1 and 2 data respectively. The results that are not available are classes that the two volunteers did not perform when collecting their data.}
\label{tab:gen}
\end{center}
\end{small}
\end{table}

\section{Conclusion}

We have demonstrated a robust and unique dataset of egocentric images that have been annotated with the user's activities, a CNN late-fusion ensemble method to classify the data, promising results in fine-tuning the model to other users and a trained model that performs well on egocentric daily living imagery. We have shown state-of-the-art results on the data compared to commonly-used methods (a traditional CNN and a Classic Ensemble) and we have determined the amount of data that is needed to train an initial CNN classifier for this problem and the amount of data that is required to fine-tune the model on a per-user basis.

\section{Acknowledgments}

This work was supported by the Intel Science and Technology Center for Pervasive Computing (ISTC-PC), and by the National Institutes of Health under award 1U54EB020404-01.

\balance

\bibliographystyle{acm-sigchi}
\bibliography{bibliography,ethomaz-bib}

\begin{thebibliography}{10}

\bibitem{Biagioni:2013tg}
Biagioni, J., and Krumm, J.
\newblock {Days of Our Lives: Assessing Day Similarity from Location Traces}.
\newblock {\em User Modeling\/} (2013).

\bibitem{Blanke:2009tq}
Blanke, U., and Schiele, B.
\newblock {Daily routine recognition through activity spotting}.
\newblock {\em Location and Context Awareness (LoCA)\/} (2009), 192--206.

\bibitem{breiman2001random}
Breiman, L.
\newblock Random forests.
\newblock {\em Machine learning 45}, 1 (2001), 5--32.

\bibitem{Clarkson:2002vh}
Clarkson, B. P.~.
\newblock {Life patterns : structure from wearable sensors}.
\newblock {\em Thesis (Ph. D.) MIT\/} (2005).

\bibitem{cover1967nearest}
Cover, T., and Hart, P.
\newblock Nearest neighbor pattern classification.
\newblock {\em IEEE Transactions on Information Theory 13}, 1 (1967), 21--27.

\bibitem{deng2009imagenet}
Deng, J., Dong, W., Socher, R., Li, L.-J., Li, K., and Fei-Fei, L.
\newblock Imagenet: A large-scale hierarchical image database.
\newblock In {\em CVPR}, IEEE (2009), 248--255.

\bibitem{Eagle:2006vw}
Eagle, N., and Pentland, A.
\newblock {Reality mining: sensing complex social systems}.
\newblock {\em Personal and Ubiquitous Computing 10}, 4 (2006), 255--268.

\bibitem{Eagle:2009jx}
Eagle, N., and Pentland, A.~S.
\newblock {Eigenbehaviors: identifying structure in routine}.
\newblock {\em Behavioral Ecology and Sociobiology 63}, 7 (2009), 1057--1066.

\bibitem{fathi2011understanding}
Fathi, A., Farhadi, A., and Rehg, J.~M.
\newblock Understanding egocentric activities.
\newblock In {\em ICCV}, IEEE (2011), 407--414.

\bibitem{hinton2012improving}
Hinton, G.~E., Srivastava, N., Krizhevsky, A., Sutskever, I., and
  Salakhutdinov, R.~R.
\newblock Improving neural networks by preventing co-adaptation of feature
  detectors.
\newblock {\em CoRR\/} (2012).

\bibitem{Hodges:2006uj}
Hodges, S., Williams, L., Berry, E., Izadi, S., Srinivasan, J., Butler, A.,
  Smyth, G., Kapur, N., and Wood, K.
\newblock Sensecam: A retrospective memory aid.
\newblock In {\em UbiComp 2006}. Springer, 2006, 177--193.

\bibitem{Hoyle:2014dj}
Hoyle, R., Templeman, R., Armes, S., Anthony, D., Crandall, D., and Kapadia, A.
\newblock {Privacy behaviors of lifeloggers using wearable cameras}.
\newblock In {\em ACM International Joint Conference} (2014), 571--582.

\bibitem{Huynh:2008tl}
Hu{\`{y}}nh, T., Fritz, M., and Schiele, B.
\newblock {Discovery of activity patterns using topic models}.
\newblock {\em Ubicomp\/} (2008).

\bibitem{jia2014caffe}
Jia, Y., Shelhamer, E., Donahue, J., Karayev, S., Long, J., Girshick, R.,
  Guadarrama, S., and Darrell, T.
\newblock Caffe: Convolutional architecture for fast feature embedding.
\newblock In {\em ACM Multimedia} (2014), 675--678.

\bibitem{Kelly:2011ei}
Kelly, P., Doherty, A., Berry, E., Hodges, S., Batterham, A.~M., and Foster, C.
\newblock {Can we use digital life-log images to investigate active and
  sedentary travel behaviour? Results from a pilot study}.
\newblock {\em International Journal of Behavioral Nutrition and Physical
  Activity 8}, 1 (May 2011), 44.

\bibitem{Kelly:2013ee}
Kelly, P., Marshall, S.~J., Badland, H., Kerr, J., Oliver, M., Doherty, A.~R.,
  and Foster, C.
\newblock {An ethical framework for automated, wearable cameras in health
  behavior research.}
\newblock {\em American journal of preventive medicine 44}, 3 (Mar. 2013),
  314--319.

\bibitem{Kerr:2013dc}
Kerr, J., Marshall, S.~J., Godbole, S., Chen, J., and Legge, A.
\newblock {Using the SenseCam to Improve Classifications of Sedentary Behavior
  in Free-Living Settings}.

\bibitem{krizhevsky2012imagenet}
Krizhevsky, A., Sutskever, I., and Hinton, G.~E.
\newblock Imagenet classification with deep convolutional neural networks.
\newblock In {\em NIPS} (2012), 1097--1105.

\bibitem{lecun1998gradient}
LeCun, Y., Bottou, L., Bengio, Y., and Haffner, P.
\newblock Gradient-based learning applied to document recognition.
\newblock {\em IEEE 86}, 11 (1998), 2278--2324.

\bibitem{Marcu:2012wz}
Marcu, G., Dey, A.~K., and Kiesler, S.
\newblock {Parent-driven use of wearable cameras for autism support: A field
  study with families}.
\newblock {\em Ubicomp 2012\/} (2012), 401--410.

\bibitem{Nguyen:2009tv}
Nguyen, D.~H., Marcu, G., Hayes, G.~R., Truong, K.~N., Scott, J., Langheinrich,
  M., and Roduner, C.
\newblock {Encountering SenseCam: personal recording technologies in everyday
  life}.
\newblock 165--174.

\bibitem{oliva2001modeling}
Oliva, A., and Torralba, A.
\newblock Modeling the shape of the scene: A holistic representation of the
  spatial envelope.
\newblock {\em IJCV 42}, 3 (2001), 145--175.

\bibitem{OLoughlin:2013ic}
O'Loughlin, G., Cullen, S.~J., McGoldrick, A., O'Connor, S., Blain, R.,
  O'Malley, S., and Warrington, G.~D.
\newblock {Using a wearable camera to increase the accuracy of dietary
  analysis.}
\newblock {\em American journal of preventive medicine 44}, 3 (Mar. 2013),
  297--301.

\bibitem{pirsiavash2012detecting}
Pirsiavash, H., and Ramanan, D.
\newblock Detecting activities of daily living in first-person camera views.
\newblock In {\em CVPR}, IEEE (2012), 2847--2854.

\bibitem{Sun:2014jk}
Sun, F.-T., Yeh, Y.-T., Cheng, H.-T., Kuo, C., and Griss, M.~L.
\newblock {Nonparametric discovery of human routines from sensor data.}
\newblock {\em PerCom\/} (2014), 11--19.

\bibitem{Sun:2010gc}
Sun, M., Fernstrom, J.~D., Jia, W., Hackworth, S.~A., Yao, N., Li, Y., Li, C.,
  Fernstrom, M.~H., and Sclabassi, R.~J.
\newblock {A wearable electronic system for objective dietary assessment}.
\newblock {\em Journal of the American Dietetic Association 110}, 1 (2010), 45.

\bibitem{Thomaz:2013iv}
Thomaz, E., Parnami, A., Bidwell, J., Essa, I.~A., and Abowd, G.~D.
\newblock {Technological approaches for addressing privacy concerns when
  recognizing eating behaviors with wearable cameras.}
\newblock {\em UbiComp\/} (2013), 739--748.

\bibitem{Willett:2002jy}
Willett, W.~C.
\newblock {Balancing Life-Style and Genomics Research for Disease Prevention}.
\newblock {\em Science 296}, 5568 (Apr. 2002), 695--698.

\bibitem{zeiler2014visualizing}
Zeiler, M.~D., and Fergus, R.
\newblock Visualizing and understanding convolutional networks.
\newblock In {\em ECCV}. Springer, 2014, 818--833.

\bibitem{Zhang:2010ki}
Zhang, H., Li, L., Jia, W., Fernstrom, J.~D., Sclabassi, R.~J., and Sun, M.
\newblock {Recognizing physical activity from ego-motion of a camera.}
\newblock {\em IEEE EMBS\/} (2010), 5569--5572.

\end{thebibliography}
\end{document}